\documentclass{article}
\usepackage{ijcai21}
\usepackage{amsfonts}

\usepackage{times}

\usepackage{soul}
\usepackage{url}
\usepackage[hidelinks]{hyperref}
\usepackage[utf8]{inputenc}
\usepackage[small]{caption}
\usepackage{graphicx}
\usepackage{amsmath}
\usepackage{booktabs}
\usepackage{graphicx}
\usepackage{subfigure}
\usepackage{multirow}

\title{Self Training with Ensemble of Teacher Models }

\author{
Soumyadeep Ghosh\footnote{Contact Author}\and
Sanjay Kumar\and
Janu Verma\And
Awanish Kumar\\
\affiliations
AI Garage, Mastercard\\

\emails
\{soumyadeep.ghosh, sanjay.kumar3, janu.verma, awanish.kumar\}@mastercard.com
}

\begin{document}

\maketitle

\begin{abstract}
  In order to train robust deep learning models, large amounts of labelled data is required. However, in the absence of such large repositories of labelled data, unlabeled data can be exploited for the same. Semi-Supervised learning aims to utilize such unlabeled data for training classification models. Recent progress of self-training based approaches have shown promise in this area, which leads to this study where we utilize an ensemble approach for the same. A by-product of any semi-supervised approach may be loss of calibration of the trained model especially in scenarios where unlabeled data may contain out-of-distribution samples, which leads to this investigation on how to adapt to such effects. Our proposed algorithm carefully avoids common pitfalls in utilizing unlabeled data and leads to a more accurate and calibrated supervised model compared to vanilla self-training based student-teacher algorithms. We perform several experiments on the popular STL-10 database followed by an extensive analysis of our approach and study its effects on model accuracy and calibration. 
\end{abstract}

\section{Introduction}
\label{intro}

Recently deep learning models have gained significant attention due to their excellent representational capacity and ground breaking results for problems such as object recognition \cite{tan2019efficientnet,han2020ghostnet}, object detection \cite{he2017mask}, image segmentation \cite{hu2018videomatch}, speech recognition \cite{ravanelli2020multi}, natural language processing \cite{zhang2020semantics}, machine translation \cite{zhu2020incorporating}, autonomous driving \cite{wang2019pseudo} and so on. In order to train such state-of-the-art models, requirement of large amounts of labelled data is inevitable. However, labelling data is an expensive and time-consuming task and requires extensive human intervention. Thus, availability of labelled data is limited. On the other hand, for most problems, abundant quantity of unlabeled data is not difficult to procure. Several advances in alternative machine learning paradigms which allow models to leverage unlabeled data such as self-supervised learning \cite{doersch2015unsupervised} and semi supervised learning \cite{zhu2009introduction} have been thoroughly investigated. Several other techniques that fall under the umbrella of weakly supervised learning \cite{zhou2018brief} have made attempts to tackle the scope of utilizing unlabeled data for training machine learning models. In addition to that most of the modern applications of deep learning have been deployed at industrial scale and need to exhibit consistent performance. As an example, for an autonomous driving system to be deemed fit for deployment, an exhaustive test of its calibration \cite{guo2017calibration} needs to be performed. 
\begin{figure}[t]
	\begin{center}
		\includegraphics[width=1\linewidth]{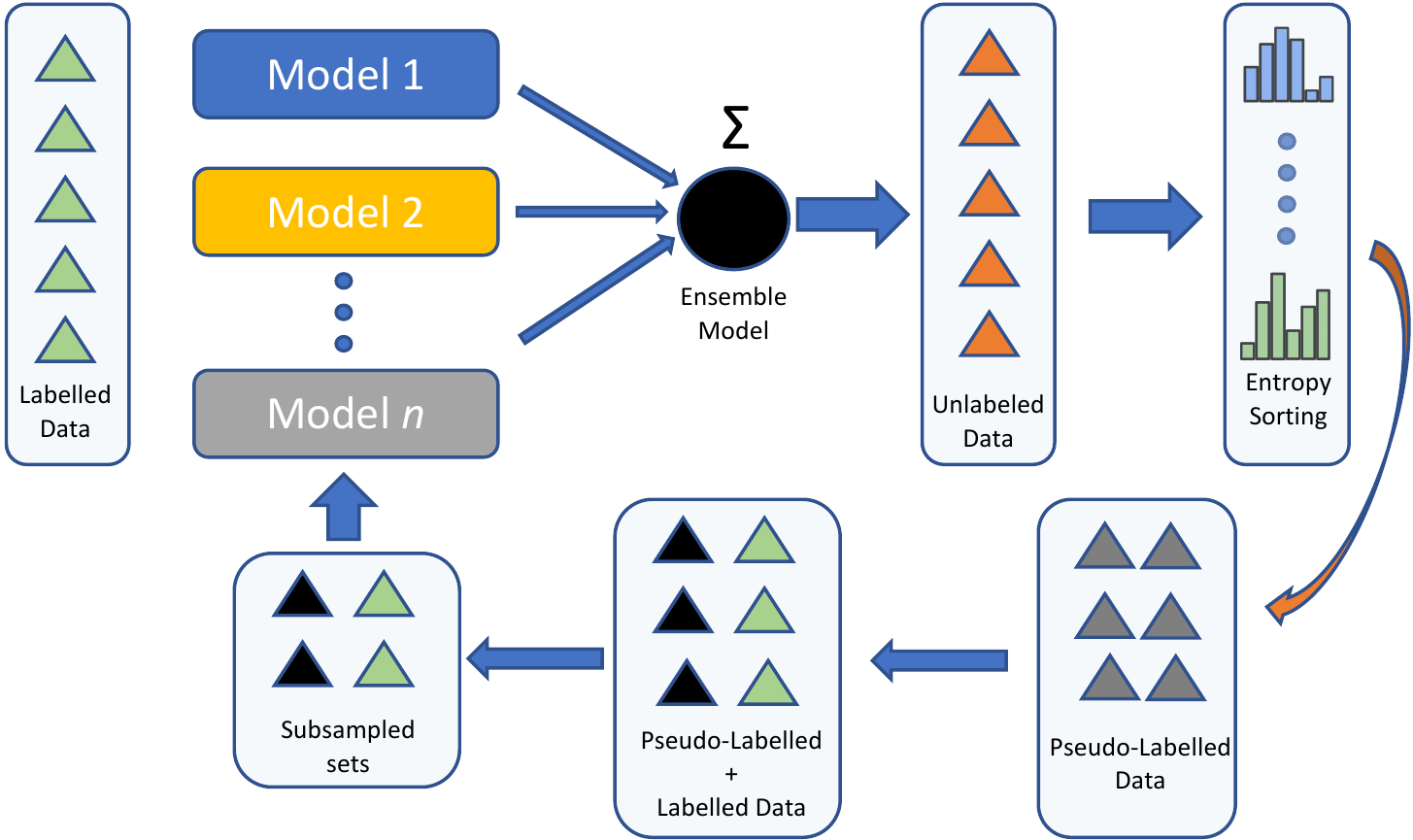}
	\end{center}
	\caption{Illustrating the concept of the proposed approach. An ensemble of teacher models are trained on different subsamples of the training data, which are then utilized during pseudolabelling of unlabelled data.}
   \vspace{-12pt} 
	\label{fig:grap_abs}
\end{figure}

Semi-Supervised approaches are focused at utilizing vast amounts of unlabeled data to improve the models trained on limited labelled data. Basic approaches of semi supervised learning use a well known technique known as pseudo-labelling~\cite{lee2013pseudo}, where the trained model is utilized to infer labels on the unlabeled data and incorporate them into the labelled set for training a  model. This expands the labeled set of data and hence in most cases leads to a better trained supervised model. Some other popular approaches for semi-supervised learning include transduction \cite{shi2018transductive}, label propagation \cite{iscen2019label} and consistency regularization \cite{tarvainen2017mean,laine2016temporal,sajjadi2016regularization}. All these approaches depend on the supervised model trained on the labelled data, which guide the process of utilization of unlabeled data. There have been several variants of the Pseudo-labelling approach, depending on the process of utilizing the pseudo-labelled data. Self-training \cite{xie2020self} with student-teacher models use a trained model (teacher) to assign labels to unlabeled data and then re-train another model (student model) on the labelled and pseudo-labelled data, which is repeated for a few iterations. In this paper we pay special focus on this popular semi-supervised learning approach and design several experiments to investigate the effect of using an ensemble of teacher models on semi supervised learning and model calibration. Such observations would also lead us to understanding the balance between incorporation of unlabeled data and model calibration. 

Pseudo-labelling \cite{lee2013pseudo} has always been one of the most convenient approaches of incorporating unlabeled data in learning a supervised model, especially for the simplicity of implementation and its effectiveness in yielding encouraging results. Another interesting observation that has been made is that pseudo-labelling induces an entropy minimization \cite{NIPS2004_96f2b50b} effect on the model \cite{NEURIPS2019_1cd138d0}. The reason for that is, since it selects data on which the model is most confident, such data points lie away from the decision boundary and hence entropy minimization comes into play when we train with such data points. 
However, it has been agreed on by several prominent works in this area, that pseudo-labelling may in some cases incorporate incorrectly labelled data. The vanilla pseudo-labelling technique uses a trained network to evaluate the unlabeled data using the highest confidence score on the unlabeled samples. However, in some cases it might happen that the model may result in incorrect inference with high confidence. In addition to this poor calibration of networks \cite{guo2017calibration} may also result in such incorrect labelling. To mitigate this effect, we utilize the self-training approach and carefully select a sample of unlabeled data instead of taking the entire set. This allows us to do away with samples which might include label noise and out-of-distribution data, although our study is mostly empirical and we do not provide any formal proof for the same. The salient contributions of our work are as follows:

\begin{itemize}
    \item We design a framework with an ensemble of teacher models where iterative label improvement is performed. We show the effect of carefully selecting only a sample of unlabeled data and increasing the size of this sample on model accuracy. 
    
    \item We also exhibit a comparative analysis of model calibration on several variants of our approach. We show that the ensemble approach of self-training performs better in terms of model calibration compared to vanilla self-training  for semi-supervised learning. 
\end{itemize}

We perform experiments on one of the most popular databases for semi-supervised learning, the STL-10 \cite{coates2011analysis} database and follow up with an extensive set of analysis for the same. 

\section{Problem Formulation}
We consider a collection of $n$ examples $X \colon =\{x_1,x_2, \ldots, x_l,x_{l+1}, \ldots, x_n\}$ with $x_i \in \mathbf{\chi}$. The first $l$ examples $X_L \colon =\{x_1,x_2, \ldots, x_l\}$ are labeled as $Y_L \colon = \{y_1, y_2, \ldots, y_l\}$ where $y_i \in C$ a discrete set over $c$ classes i.e. $C\colon= \{1,2, \ldots, c\}$. The remaining examples $x_i$ for $i \in U \colon= \{l+1, l+2, \ldots, n \}$ are unlabeled. Denote by $X_U$ the unlabeled set, then $X$ is the disjoint union of the two sets $X = X_L \cup X_U$. In supervised learning, we use the labeled examples with their corresponding labels $(X_L, Y_L)$ to train a classifier that learns to predict class-labels for previously unseen examples. The guiding principle of semi-supervised learning is to leverage the unlabeled examples as well to train the classifier. 
\\
\\
{\bf Supervised Learning:} We assume a deep convolutional neural network (DCNN) based classifier trained on the labeled set of examples $(X_L, Y_L)$ which takes an example $x_i \in \chi$ and outputs a vector of class-label probabilities i.e. $f_{\theta} : \chi \rightarrow \mathbb{R}^c$ where $\theta$ are the parameters of the model. The model is trained by minimizing the {\em supervised loss} -
\begin{equation}
\label{supervised}
    L_s (X_L, Y_L, \theta) = \sum_{i=1}^l l_s(f_{\theta}(x_i), y_i)
\end{equation}
A typical choice for the loss function $l_s$ in classification is the {\em cross-entropy} $l_s(\hat{y}, y) = - y \text{log}(\hat{y})$ \\
The DCNN can be thought of as the composition of two networks - {\em feature extraction network} which transforms an input example to a vector of features $\phi_{\theta} : \chi \rightarrow \mathbb{R}^d$ and {\em classification network} which maps the feature vector to the class vector. Let $v_i \colon= \phi_{\theta}(x_i)$ be the feature vector of $x_i$. The classification network is usually a {\em fully-connected layer} on top of $\phi_{\theta}$. The output of the network for $x_i$ is $f_{\theta}(x_i)$ and the final prediction is the class with highest probability score i.e 
\begin{equation}
\label{prediction}
    \hat{y}_i:= \text{arg max}_j (f_{\theta}(x_i))_j
\end{equation}
. A trained classifier (at least the feature generator network) is the starting point of the most of the semi-supervised learning techniques, including the studies performed in this work.  
\\
\\
{\bf Semi-supervised Learning (SSL):} There are two main schools of SSL approaches for image classification 
\begin{itemize}
    \item {\em Consistency Regularization:} An additional loss term called {\em unsupervised-loss} is added for either all images or for only unlabeled ones which encourages consistency under various transformations of the data.
    \begin{equation}
        L_u(X; \theta) = \sum_{i=1}^n l_u(f_{\theta}(x_i), f_{\theta}(\Tilde{x_i}))
    \end{equation}
    where $\Tilde{x_i}$ is a transformation of $x_i$. A choice for consistency loss is is the squared Euclidean distance. 
    \item {\em Pseudo-labeling:} The unlabeled examples are assigned pseudo-labels thereby expanding the label set to all of $X$. A model is then trained on this labeled set $(X_L \cup X_U), (Y_L \cup \hat{Y}_U)$ using the supervised loss for the true-labeled examples plus a similar loss for the pseudo-labeled examples. 
    \begin{equation}
         L_p(X_U, \hat{Y}_U, \theta) = \sum_{i=l+1}^n l_s(f_{\theta}(x_i), \hat{y}_i)
    \end{equation}
\end{itemize}
The current work fits in the realm of the later school where we study the effect of iteratively adding pseudo-labeled examples for self-training.\\ 
\\
\textbf{Self Training using Student-Teacher Models: }
This class of methods \cite{xie2020self} for SSL iteratively use a trained (teacher) model to pseudo-label a set of unlabeled examples, and then re-train the model (now student) on the labelled plus the pseudo-labelled examples. Usually the same model assumes the dual role of the the student (as the learner) and the teacher (it generates labels, which are then used by itself as a student for learning). A model $f_{\theta}$ is trained on the labelled data $X_L$ (using supervised loss equation~\ref{supervised}), and is then employed for inference on the unlabeled set $X_U$. The prediction vectors $f_{\theta}(x_i) \forall x_i \in X'_U$ are converted to one-hot-vectors, where $X'_U \subset X_U$. These examples $X'_U$ along with their corresponding (pseudo-)labels $\hat{Y}'_U$ are added to the original labelled set. This extended labelled set $X_L \cup X'_U$ is used to train another (student) model $f'_{\theta}$. This procedure is repeated and the current student model is used as a teacher in the next phase to get pseudo-labels $\cup X''_U$ for training another (student) model $f''_{\theta}$ on the set $X_L \cup X'_U \cup X''_U$. Now, for conventional self-training methods we use the entire unlabeled set $X_U$ in every iteration. However, as mentioned above, the most general form of self training can have different sets of unlabeled data ($X'_U$, $X''_U$ and so on) in every iteration. The method of selection of $X'_U$ from $X_U$ can come from any utility function, the objective of which would be to use the most appropriate unlabeled data samples in each iteration. Some methods even use weights for each (labelled/unlabeled) data sample, which are updated in every iteration, similar to a process followed in Transductive Semi-Supervised Learning~\cite{shi2018transductive} methods, which is borrowed from the traditional concept of boosting used in statistics. 
\\
\\
\textbf{Calibration and Robustness:} It is desirable that a classifier should provide a calibrated uncertainty measure in addition to its prediction accuracy. The concept of  model calibration stems from the original concept of how probabilistically correct a classifier is \cite{niculescu2005predicting}, on unseen test data.  A classifier is {\em well-calibrated}, if the probability associated with the predicted class label matches the probability of such prediction being correct. Recently several studies \cite{niculescu2005predicting,guo2017calibration,mozafari2019unsupervised,kull2019beyond} have illustrated its importance and demonstrated that deep neural network based classifiers are not well-calibrated. Such models often give overconfident predictions, which are evident by the observation that their average correctness (accuracy) ($A_c$) on unseen test data is far inferior than the average maximum prediction probability for the most probable class. We study the calibration of a classifier in terms of its calibration error. \\
For a classifier $f_{\theta}$,  the {\em average maximum prediction probability} is defined as 
\begin{equation}
    \hat{s}_i:=   1/n\sum_{i=1}^{n}\text{max}(f_{\theta}(x_i)) 
\end{equation}
where $n$ is the total numbers of test samples. \\
The {\em calibration error} of the classifier $f_{\theta}$  is defined as 
\begin{equation}
    \mathcal{E}=A_c - \hat{s}_i    
\end{equation}
The lower the value of $\mathcal{E}$, the better-calibrated the model is. \\
An additional goal for the current study is to investigate the effect of pseudo-labeling on model robustness which will be measure via its calibration error. Contemporary applications such as self driving cars, video surveillance systems etc. are high-regret situations and require well calibrated models.

\section{Method}
In the following, we describe our recipe for iterative pseudo-labeling based semi-supervised learning. In nutshell, we iteratively pseudo-label the unlabeled examples and use a set of high-confidence pseudo-labeled examples for re-training the model. There are 3 main ingredients - subsampling, training, pseudo-labeling. These are discussed below. The graphical overview of the approach is shown in Figure \ref{fig:grap_abs}. 
\\
\\
{\bf Sub-sampling} Let $T^{(i)}=(X^{(i)},Y^{(i)})$ be a set of training examples where each $x_i \in X^{(i)}$ has a corresponding label $y_i \in Y^{(i)}$. We generate $k$ random samples of the training set $T^{(i)}$ each of size $m$, $m < |T^{(i)}|$. An example can be present in more than one of the $k$ samples i.e. we do not require $m * k$ to be equal to  $|T^{(i)}|$.  
\\
\\
{\bf Model Training} We train $k$ separate models on each of the $k$ samples of the training data. The models are also chosen to be different architectures with their separate parameters $\theta_1, \theta_2, \ldots, \theta_k$. The unlabeled examples $X_U$ are then fed to each of the $k$ trained models to infer their corresponding probability vectors. We obtain $(f_{\theta_1}(x), f_{\theta_2}(x), \ldots, f_{\theta_k}(x))$ for each $x \in X_U$, and  $f_{\theta_i}(x) \in \mathbb{R}^c \forall i$.  
\\
\\
{\bf Pseudo-labeling} For assigning pseudo-labels to the unlabeled examples, we take the {\em ensemble} of the predictions of the individual $k$ models. 
\begin{equation}
    f_{ensmbl}(x) = \frac{1}{k}\sum_{j=1}^k f_{\theta_j}(x)
\end{equation}
The unlabeled examples are then sorted in decreasing order of the entropy of the ensemble prediction vectors $f_{ensmbl}(x)$, and first $p$ examples with the lowest entropy are selected.
We assign pseudo-labels $\hat{y}$ to these $p$ examples as follows:
\begin{equation}
    \hat{y}_i = \text{arg max}_j (f_{ensmbl}(x_i))_j
\end{equation}
The rationale for entropy sorting is that having lower entropy prediction vector implies the model being more confident at those examples. In some SSL approaches (e.g. label propagation) use entropy, as a measure of uncertainty, to assign weight to the pseudo-labels. In that sense, selecting top $p$ examples translates to having same weight equal to 1 for first $p$ examples and weight equal to 0 for all the others.These $p$ examples with their corresponding pseudo-labels are added to the training set. \\
The procedure we follow is outlined in the following steps:
\begin{enumerate}
    \item Let $T_0=(X^0,Y_0) = (X_L,Y_L)$ be the initial training data.
    \item for $i=0$ to $i=N$, perform the following steps:
        \begin{enumerate}
            \item Create $k$ random sample of $T^{(i)}$ as described above. 
            \item Train a separate model on each of the $k$ samples.
            \item For each $x \in X_U$, get its prediction vector from each of the $k$ models. $(f_{\theta_1}(x), f_{\theta_2}(x), \ldots, f_{\theta_k}(x))$.
            \item Compute the ensemble probability vector $f_{ensmbl}(x) \forall x \in X_U$. 
            \item Sort the unlabeled examples in terms of entropy, and choose first $p$ examples $X_p \subset X_U$. 
            \item Assign one-hot-encoded pseudo-labels $\hat{Y}_p$ for all $x \in X_p$. 
            \item Create the training data for next iteration as :  $T^{(i+1)}=(X^{(i+1)},Y^{(i+1)}) = (X^{(i)} \cup X_p,Y^{(i)} \cup \hat{Y}_p)$
        \end{enumerate}
\end{enumerate}

\begin{table*}[]
\centering
\small
\caption{Classification accuracy (.xyz to be read as xy.z \%) on the STL-10 database. Experiment 1 shows results where we do not use ensemble, experiments 2 and 3 utilized ensembles without and with subsampling respectively}
\begin{tabular}{|c|l|c|c|c|c|}
\hline
\textbf{Experiment} & \multicolumn{1}{c|}{\textbf{Model}} & \textbf{\begin{tabular}[c]{@{}c@{}}Base Iteration\\ (5k Labelled)\end{tabular}} & \textbf{\begin{tabular}[c]{@{}c@{}}Iteration 1\\ (5k + 10k)\end{tabular}} & \textbf{\begin{tabular}[c]{@{}c@{}}Iteration 2\\ (5k + 20k)\end{tabular}} & \textbf{\begin{tabular}[c]{@{}c@{}}Iteration 3\\ (5k + 30k)\end{tabular}} \\ \hline
\multirow{3}{*}{1}  & ResNet 18                           & 0.7221                                                                          & 0.7253                                                                    & 0.7316                                                                    & 0.7412                                                                    \\ \cline{2-6} 
                    & Efficient-Net B0                    & 0.6721                                                                           & 0.7212                                                                    & 0.7392                                                                    & 0.7365                                                                    \\ \cline{2-6} 
                    & Wide-Resnet 50                      & 0.6365                                                                          & 0.6855                                                                    & 0.7008                                                                    & 0.7038                                                                    \\ \hline
\multirow{4}{*}{2}  & ResNet 18                           & 0.7221                                                                          & 0.7326                                                                    & 0.7523                                                                    & 0.7554                                                                    \\ \cline{2-6} 
                    & Efficient-Net B0                    & 0.6721                                                                           & 0.7387                                                                    & 0.7478                                                                    & 0.7572                                                                    \\ \cline{2-6} 
                    & Wide-Resnet 50                      & 0.6365                                                                          & 0.7092                                                                    & 0.7184                                                                    & 0.7496                                                                    \\ \cline{2-6} 
                    & Ensemble (Without Subsampling)                           & 0.7045                                                                          & 0.7651                                                                    & 0.7691                                                                    & 0.7743                                                                    \\ \hline
\multirow{4}{*}{3}  & ResNet 18                           & 0.6808                                                                          & 0.7296                                                                    & 0.7542                                                                    & 0.7648                                                                    \\ \cline{2-6} 
                    & Efficient-Net B0                    & 0.652                                                                           & 0.7423                                                                    & 0.7515                                                                    & 0.7688                                                                    \\ \cline{2-6} 
                    & Wide-Resnet 50                      & 0.6213                                                                          & 0.7143                                                                    & 0.74725                                                                   & 0.7581                                                                    \\ \cline{2-6} 
                    & Ensemble  (With Subsampling)                          & 0.7045                                                                          & \textbf{0.7672}                                                           & \textbf{0.7855}                                                           & \textbf{0.7888}                                                           \\ \hline
\end{tabular}
\end{table*}

\subsection{Relation to Transductive SSL} A simple technique for pseudo-labeling based SSL is designed by \cite{lee2013pseudo} where they first train a network $f_{\theta}$ on the labeled examples $X_L$ and then assign pseudo-labels for the unlabeled examples according to equation \ref{prediction} for $x_i \in X_U$. A limitation of this approach is that it ignores the variation in the degree of uncertainty in the pseudo-labels for the unlabeled examples - all the pseudo-labelled examples are treated as the same and are all used in the model training. Some methods e.g. developed by \cite{shi2018transductive} use uncertainty weights in the loss-function during training on the pseudo-labelled examples. In the current work, we only use a set of high-confidence pseudo-labelled examples for training the model. Like \cite{shi2018transductive}, we follow an iterative procedure of pseudo-labeling and model training. Our approach, however, uses only a set of pseudo-labelled examples at each step of the iteration. Moreover, the model at each step of the iteration is an ensemble of $k$ models each trained on a separate sample of the data.

\subsection{Relation to Teacher-Student Models} 
Our approach fits in the teacher-student paradigm since we are following the same philosophy of iterating over generating pseudo-labels (teacher) and re-training the model (student). The main difference is that we are using an ensemble of the models to pseudo-label the unlabeled examples and at each iteration the pseudo-labelled examples are used by $k$ different models for training. Following the analogy, as a teacher the model aggregates from multiple models and dissipates the learnings to $k$ different students. \\
\\
There are some variants of the teacher-student framework e.g. noisy teacher-student \cite{xie2020self}, and mean-teacher \cite{tarvainen2017mean} that have used ensembling for semi supervised learning. The utility of self-training has been exhibited in these studies, however there are significant differences of such approaches compared to this study. These approaches do not explore  iterative ensembling and using only a subset of unlabeled data in each iteration. The current study also uses a greedy incremental selection of unlabeled data using the entropy of the prediction vector obtained from them, which makes it different from other related work done in this domain. 


\section{Experiments}
This section outlines the experiments followed by analysis and ablation studies for additional insights and discussion. 
\subsection{Databases}
We perform experiments on the \textbf{STL-10}~\cite{coates2011analysis} database which contains 113,000 images (of resolution $96 \times 96$) of 10 object categories of resolution $96 \times 96$. The training set contains 5,000 images and 8,000 images are in the testing set. Each class has 500 images for training and 800 images for testing. The rest of the images are unlabeled which include images from a different distribution (classes other than those which are there in the training set). 

\subsection{Experimental Protocol}
Our study is focused on the impact of ensembling of teacher models on semi-supervised learning and model calibration. In order to present this investigation in a structured fashion we present three primary experiments which are illustrated as follows:

\begin{enumerate}
  
   \item \textbf{Ensemble with Subsampling} In order to create the ensembles we randomly select 4000 samples out of the 5000 labelled samples in the database. We repeat this process and prepare 3 such samples containing 4000 samples in each set. This allows repetition of a data sample in more than one set. We train 3 separate models with different architectures and set of parameters on these 3 sets - each model is trained on 4000 data examples. Once these models are trained we use an ensemble of these models to pseudo-label the unlabeled data. In the next iteration again 3 models will be trained on the labelled plus pseudo-labelled examples. This is essentially an iterative boosting-like method where each iteration is similar to a bagging-like approach. 
   
   \item \textbf{Ensemble without Subsampling:} This experiment is similar to the above, the only difference being that we do not subsample a smaller set for training different models. The 3 separate models are trained on all of the labelled and pseudo-labelled examples in each iteration. For example, in the first iteration we use all of the 5000 samples to train the models, and then the ensemble is used to pseudo-label the unlabeled examples. This experiment helps us understand the effect of sub-sampling while training an ensemble of teacher models. 
   
   \item \textbf{Non-Ensemble Approach:} In this experiment we use a conventional self-training paradigm where only 1 teacher model is used and that is used to pseudo-label the unlabeled data in every iteration. This experiment allows us to observe the benefits of our ensembling approach. 
\end{enumerate}

\begin{table*}[t]
\centering
\small
\caption{Calibration error (.xyz to be read as xy.z \%) on the STL-10 database. Experiment 1 shows results where we do not use ensemble, experiments 2 and 3 utilized ensembles without and with subsampling respectively}
\begin{tabular}{|c|l|c|c|c|c|}
\hline
\textbf{Experiment} & \multicolumn{1}{c|}{\textbf{Model}} & \textbf{\begin{tabular}[c]{@{}c@{}}Base Iteration\\ (5k Labelled)\end{tabular}} & \textbf{\begin{tabular}[c]{@{}c@{}}Iteration 1\\ (5k + 10k)\end{tabular}} & \textbf{\begin{tabular}[c]{@{}c@{}}Iteration 2\\ (5k + 20k)\end{tabular}} & \textbf{\begin{tabular}[c]{@{}c@{}}Iteration 3\\ (5k + 30k)\end{tabular}} \\ \hline
\multirow{3}{*}{1}  & ResNet 18                           & 0.134                                                                           & 0.145                                                                     & 0.157                                                                     & 0.166                                                                     \\ \cline{2-6} 
                    & Efficient-Net B0                    & 0.050                                                                           & 0.079                                                                     & 0.114                                                                     & 0.094                                                                     \\ \cline{2-6} 
                    & Wide-Resnet 50                      & 0.114                                                                           & 0.155                                                                     & 0.158                                                                     & 0.168                                                                     \\ \hline
\multirow{4}{*}{2}  & ResNet 18                           & 0.134                                                                           & 0.140                                                                     & 0.150                                                                     & 0.163                                                                     \\ \cline{2-6} 
                    & Efficient-Net B0                    & 0.050                                                                           & 0.092                                                                     & 0.089                                                                     & 0.120                                                                     \\ \cline{2-6} 
                    & Wide-Resnet 50                      & 0.114                                                                           & 0.135                                                                     & 0.139                                                                     & 0.148                                                                     \\ \cline{2-6} 
                    & Ensemble (Without Subsampling)                            & 0.018                                                                          & 0.025                                                                     & 0.042                                                                     & 0.076                                                                     \\ \hline
\multirow{4}{*}{3}  & ResNet 18                           & 0.126                                                                           & 0.142                                                                     & 0.149                                                                     & 0.136                                                                     \\ \cline{2-6} 
                    & Efficient-Net B0                    & 0.098                                                                           & 0.066                                                                     & 0.090                                                                     & 0.105                                                                     \\ \cline{2-6} 
                    & Wide-Resnet 50                      & 0.094                                                                           & 0.122                                                                     & 0.127                                                                     & 0.138                                                                     \\ \cline{2-6} 
                    & Ensemble (With Subsampling)                           & -0.013                                                                          & \textbf{0.013}                                                            & \textbf{0.032}                                                            & \textbf{0.051}                                                            \\ \hline
\end{tabular}
\end{table*}

The above three investigations only differ in the process of ensembling, followed by a series of steps which are similar across the experiments. The steps are as follows,

\begin{itemize}
    \item \textbf{Averaging Predictions:} The constituent models are used for inference on the unlabeled data. The prediction vectors from the ensemble models are averaged which gives us one prediction vector for each unlabeled data.
    
    \item \textbf{Entropy Sorting:} In this step our objective is to use only those unlabeled data samples on which the ensemble is most confident. Such predictions are represented by low entropy on the prediction vector. So the unlabeled samples corresponding to the lowest entropy prediction vectors are chosen for training the individual models for the next iteration. 
    
    \item \textbf{Pseudo-labelling:} The prediction vector obtained in the previous step is converted into a one-hot label for the unlabeled sample. This (pseudo-)label would be used in the next iteration to train the next set of models. It should also be noted that in later iterations, previously pseudo-labelled data (from the unlabeled set) is again put through this step with the latest model available in that iteration. 
    
    \item \textbf{Combining labeled and unlabeled data:} The lowest entropy yielding unlabeled data samples are allowed to participate in the next iteration of training the ensemble models along with the labeled data. This combination would go through the sub-sampling process (in the first experiment only) and then the next steps are repeated. 
    
\end{itemize}


\subsection{Implementation Details}
The backbone of our proposed approach are 3 popular deep convolutional neural network models, namely ResNet-18 \cite{he2016deep}, EfficientNet-B0 \cite{tan2019efficientnet} and Wide-ResNet 50 \cite{zagoruyko2016wide}. These networks are used in subsequent iterations in our approach, and their structure and parameters are kept consistent throughout. On the STL-10 database, we subsample 4000 labelled samples randomly with replacement, and prepare 3 such sets. On each set we train a separate model. The ensembling process involves taking the simple average of 3 prediction vectors for each unlabeled sample. For the entropy sorting step we utilize Shannon's Entropy. For optimizing the models we use backpropagation with Adam \cite{kingma2014adam} optimizer. A learning rate of $10^{-3}$ has been used with 16 as batch size. The augmentation used in training the labelled and pseudolabelled dataset are Random Horizontal FLip with 0.25 probability, Random Vertical Flip with 0.25 probability, Random Rotation of (-20 degrees, 20 degrees), Random Horizontal shift (-0.25, 0.25), Random Vertical Shift (-0.25, 0.25) and random shear (-0.25, 0.25).  We have used the Pytorch \cite{paszke2019pytorch} library for all the three models. Our codes have been run on a system with Intel Xeon W-1290E CPU, Quadro RTX 6000 GPU with 64GB of RAM.

While sorting based on entropy is very uncommon in semi-supervised learning literature~\cite{chapelle2009semi,zhu2009introduction}, our method is much more close to active learning, where the proposed pseudo-labelling algorithm  replaces the human annotator, which is often utilized in active learning methods~\cite{prince2004does,felder2009active}. 

\subsection{Results}
The results for the three experiments are outlined in Table 1 and Table 2. The primary observation is that, the ensemble with subsampling framework gives the best results in terms of both accuracy (Table 1) and calibration error (Table 2). The experiment with ensembling but without the subsampling step suffers from calibration issues, although the loss in accuracy of the model in the final iteration is marginal. This leads to the primary takeaway from this study, that for semi-supervised learning, an ensemble of teacher models gives good results along with keeping the calibration error in check. The three models have also been chosen to be very different from each other in terms of architecture, although they are not the state-of-the-art models available. The models have been kept relatively simple to exhibit the efficacy of the ensemble.  We present some of the salient analysis of our study in the next subsection. 

\section{Analysis of Results}
In this section we delve into the analysis of our investigation from the set of experiments performed. 
\subsection{Effect of Ensembling}
The most promising aspect of our study is the effect of ensembling of the teacher models which has a clear impact on not only the accuracy in each iteration, but also on the calibration error. When we compare this to the experiment where only individual teacher models are trained we observe that they lose out on both these parameters. This leads us to the conclusion that the ensemble of teacher models provide improved performance compared to conventional self training by individual teacher models. 
\subsection{Effect on Sample Size}
Our experiments provide useful insights into the size of the sample utilized from the unlabeled pool of data. Previous studies on self-training by student-teacher models have utilized the entire pool of the unlabeled data available. However, in our case a real world scenario is depicted where label noise (out of distribution data) occurs in the database that we have used. Moreover, using a sample of unlabeled data also allows us to study the impact of sample size on accuracy and model calibration. Due to these reasons, our approach follows the iterative label improvement strategy and we observed that with increase in sample size of unlabeled data, the accuracies on the unseen test set (which is kept constant across all the experiments and iterations) improve. However, we also observe that beyond a certain level, the accuracy improvement is marginal.   
\subsection{Model Calibration}
First observation of this study of classifier calibration is that incorporating pseudo-labels results in less-calibrated models - the calibration errors increases in each iteration when we add more pseudo-labelled data. Thus pseudo-label based paradigms like self-training, SSL, active learning etc.reduce the robustness of the models. The advantages of the ensemble based approach is evident as it provides enhanced calibration, exhibited by the lower calibration error for the ensemble compared to the individual models (Table 2). It is also noted that the sub-sampling before the ensembling helps in further reducing the calibration error. This allows us to obtain a model with not only better accuracy, but also with more probabilistic correctness.    
\section{Conclusion and Future Work}
We present en extensive study of self-training using the student-teacher paradigm where iterative label improvement is performed by repeated pseudo-labelling of unlabeled data. In addition to that we also show that entropy based selection and using a sample of unlabeled data provides a good trade off between accuracy and training time, while keeping the model calibration in check. We would also like to study the effects of using temperature scaling, soft labels, sharpening, Mixup operations etc on the ensembling based paradigm presented in this study. We plan on extending this investigation to prepare a proposed approach for semi-supervised learning which can outperform the present state-of-the art approaches, while maintaining good model calibration and robustness.

\bibliographystyle{named}
\bibliography{ijcai21}

\begin{thebibliography}{}

\bibitem[\protect\citeauthoryear{Berthelot \bgroup \em et al.\egroup
  }{2019}]{NEURIPS2019_1cd138d0}
David Berthelot, Nicholas Carlini, Ian Goodfellow, Nicolas Papernot, Avital
  Oliver, and Colin~A Raffel.
\newblock Mixmatch: A holistic approach to semi-supervised learning.
\newblock In {\em Advances in Neural Information Processing Systems},
  volume~32, 2019.

\bibitem[\protect\citeauthoryear{Chapelle \bgroup \em et al.\egroup
  }{2009}]{chapelle2009semi}
Olivier Chapelle, Bernhard Scholkopf, and Alexander Zien.
\newblock Semi-supervised learning (chapelle, o. et al., eds.; 2006)[book
  reviews].
\newblock {\em IEEE Transactions on Neural Networks}, 20(3):542--542, 2009.

\bibitem[\protect\citeauthoryear{Coates \bgroup \em et al.\egroup
  }{2011}]{coates2011analysis}
Adam Coates, Andrew Ng, and Honglak Lee.
\newblock An analysis of single-layer networks in unsupervised feature
  learning.
\newblock In {\em Proceedings of the fourteenth international conference on
  artificial intelligence and statistics}, pages 215--223. JMLR Workshop and
  Conference Proceedings, 2011.

\bibitem[\protect\citeauthoryear{Doersch \bgroup \em et al.\egroup
  }{2015}]{doersch2015unsupervised}
Carl Doersch, Abhinav Gupta, and Alexei~A Efros.
\newblock Unsupervised visual representation learning by context prediction.
\newblock In {\em IEEE International Conference on Computer Vision}, pages
  1422--1430, 2015.

\bibitem[\protect\citeauthoryear{Felder and Brent}{2009}]{felder2009active}
Richard~M Felder and Rebecca Brent.
\newblock Active learning: An introduction.
\newblock {\em ASQ higher education brief}, 2(4):1--5, 2009.

\bibitem[\protect\citeauthoryear{Grandvalet and
  Bengio}{2005}]{NIPS2004_96f2b50b}
Yves Grandvalet and Yoshua Bengio.
\newblock Semi-supervised learning by entropy minimization.
\newblock In {\em Advances in Neural Information Processing Systems},
  volume~17, 2005.

\bibitem[\protect\citeauthoryear{Guo \bgroup \em et al.\egroup
  }{2017}]{guo2017calibration}
Chuan Guo, Geoff Pleiss, Yu~Sun, and Kilian~Q Weinberger.
\newblock On calibration of modern neural networks.
\newblock In {\em International Conference on Machine Learning}, pages
  1321--1330. PMLR, 2017.

\bibitem[\protect\citeauthoryear{Han \bgroup \em et al.\egroup
  }{2020}]{han2020ghostnet}
Kai Han, Yunhe Wang, Qi~Tian, Jianyuan Guo, Chunjing Xu, and Chang Xu.
\newblock Ghostnet: More features from cheap operations.
\newblock In {\em IEEE/CVF Conference on Computer Vision and Pattern
  Recognition}, pages 1580--1589, 2020.

\bibitem[\protect\citeauthoryear{He \bgroup \em et al.\egroup
  }{2016}]{he2016deep}
Kaiming He, Xiangyu Zhang, Shaoqing Ren, and Jian Sun.
\newblock Deep residual learning for image recognition.
\newblock In {\em IEEE/CVF Conference on Computer Vision and Pattern
  Recognition}, pages 770--778, 2016.

\bibitem[\protect\citeauthoryear{He \bgroup \em et al.\egroup
  }{2017}]{he2017mask}
Kaiming He, Georgia Gkioxari, Piotr Doll{\'a}r, and Ross Girshick.
\newblock Mask r-cnn.
\newblock In {\em IEEE International Conference on Computer Vision}, pages
  2961--2969, 2017.

\bibitem[\protect\citeauthoryear{Hu \bgroup \em et al.\egroup
  }{2018}]{hu2018videomatch}
Yuan-Ting Hu, Jia-Bin Huang, and Alexander~G Schwing.
\newblock Videomatch: Matching based video object segmentation.
\newblock In {\em European Conference on Computer Vision}, pages 54--70, 2018.

\bibitem[\protect\citeauthoryear{Iscen \bgroup \em et al.\egroup
  }{2019}]{iscen2019label}
Ahmet Iscen, Giorgos Tolias, Yannis Avrithis, and Ondrej Chum.
\newblock Label propagation for deep semi-supervised learning.
\newblock In {\em IEEE/CVF Conference on Computer Vision and Pattern
  Recognition}, pages 5070--5079, 2019.

\bibitem[\protect\citeauthoryear{Kingma and Ba}{2014}]{kingma2014adam}
Diederik~P Kingma and Jimmy Ba.
\newblock Adam: A method for stochastic optimization.
\newblock {\em arXiv preprint arXiv:1412.6980}, 2014.

\bibitem[\protect\citeauthoryear{Kull \bgroup \em et al.\egroup
  }{2019}]{kull2019beyond}
Meelis Kull, Miquel Perello-Nieto, Markus K{\"a}ngsepp, Hao Song, Peter Flach,
  et~al.
\newblock Beyond temperature scaling: Obtaining well-calibrated multiclass
  probabilities with dirichlet calibration.
\newblock {\em arXiv preprint arXiv:1910.12656}, 2019.

\bibitem[\protect\citeauthoryear{Laine and Aila}{2016}]{laine2016temporal}
Samuli Laine and Timo Aila.
\newblock Temporal ensembling for semi-supervised learning.
\newblock {\em arXiv preprint arXiv:1610.02242}, 2016.

\bibitem[\protect\citeauthoryear{Lee and others}{2013}]{lee2013pseudo}
Dong-Hyun Lee et~al.
\newblock Pseudo-label: The simple and efficient semi-supervised learning
  method for deep neural networks.
\newblock In {\em ICML Workshop on Challenges in Representation Learning},
  volume~3, 2013.

\bibitem[\protect\citeauthoryear{Mozafari \bgroup \em et al.\egroup
  }{2019}]{mozafari2019unsupervised}
Azadeh~Sadat Mozafari, Hugo~Siqueira Gomes, Wilson Le{\~a}o, and Christian
  Gagn{\'e}.
\newblock Unsupervised temperature scaling: An unsupervised post-processing
  calibration method of deep networks.
\newblock {\em arXiv preprint arXiv:1905.00174}, 2019.

\bibitem[\protect\citeauthoryear{Niculescu-Mizil and
  Caruana}{2005}]{niculescu2005predicting}
Alexandru Niculescu-Mizil and Rich Caruana.
\newblock Predicting good probabilities with supervised learning.
\newblock In {\em International Conference on Machine Learning}, pages
  625--632, 2005.

\bibitem[\protect\citeauthoryear{Paszke \bgroup \em et al.\egroup
  }{2019}]{paszke2019pytorch}
Adam Paszke, Sam Gross, Francisco Massa, Adam Lerer, James Bradbury, Gregory
  Chanan, Trevor Killeen, Zeming Lin, Natalia Gimelshein, Luca Antiga, et~al.
\newblock Pytorch: An imperative style, high-performance deep learning library.
\newblock {\em arXiv preprint arXiv:1912.01703}, 2019.

\bibitem[\protect\citeauthoryear{Prince}{2004}]{prince2004does}
Michael Prince.
\newblock Does active learning work? a review of the research.
\newblock {\em Journal of Engineering Education}, 93(3):223--231, 2004.

\bibitem[\protect\citeauthoryear{Ravanelli \bgroup \em et al.\egroup
  }{2020}]{ravanelli2020multi}
Mirco Ravanelli, Jianyuan Zhong, Santiago Pascual, Pawel Swietojanski, Joao
  Monteiro, Jan Trmal, and Yoshua Bengio.
\newblock Multi-task self-supervised learning for robust speech recognition.
\newblock In {\em IEEE International Conference on Acoustics, Speech and Signal
  Processing}, pages 6989--6993, 2020.

\bibitem[\protect\citeauthoryear{Sajjadi \bgroup \em et al.\egroup
  }{2016}]{sajjadi2016regularization}
Mehdi Sajjadi, Mehran Javanmardi, and Tolga Tasdizen.
\newblock Regularization with stochastic transformations and perturbations for
  deep semi-supervised learning.
\newblock {\em arXiv preprint arXiv:1606.04586}, 2016.

\bibitem[\protect\citeauthoryear{Shi \bgroup \em et al.\egroup
  }{2018}]{shi2018transductive}
Weiwei Shi, Yihong Gong, Chris Ding, Zhiheng~MaXiaoyu Tao, and Nanning Zheng.
\newblock Transductive semi-supervised deep learning using min-max features.
\newblock In {\em European Conference on Computer Vision}, pages 299--315,
  2018.

\bibitem[\protect\citeauthoryear{Tan and Le}{2019}]{tan2019efficientnet}
Mingxing Tan and Quoc Le.
\newblock Efficientnet: Rethinking model scaling for convolutional neural
  networks.
\newblock In {\em International Conference on Machine Learning}, pages
  6105--6114, 2019.

\bibitem[\protect\citeauthoryear{Tarvainen and
  Valpola}{2017}]{tarvainen2017mean}
Antti Tarvainen and Harri Valpola.
\newblock Mean teachers are better role models: Weight-averaged consistency
  targets improve semi-supervised deep learning results.
\newblock {\em arXiv preprint arXiv:1703.01780}, 2017.

\bibitem[\protect\citeauthoryear{Wang \bgroup \em et al.\egroup
  }{2019}]{wang2019pseudo}
Yan Wang, Wei-Lun Chao, Divyansh Garg, Bharath Hariharan, Mark Campbell, and
  Kilian~Q Weinberger.
\newblock Pseudo-lidar from visual depth estimation: Bridging the gap in 3d
  object detection for autonomous driving.
\newblock In {\em IEEE/CVF Conference on Computer Vision and Pattern
  Recognition}, pages 8445--8453, 2019.

\bibitem[\protect\citeauthoryear{Xie \bgroup \em et al.\egroup
  }{2020}]{xie2020self}
Qizhe Xie, Minh-Thang Luong, Eduard Hovy, and Quoc~V Le.
\newblock Self-training with noisy student improves imagenet classification.
\newblock In {\em IEEE/CVF Conference on Computer Vision and Pattern
  Recognition}, pages 10687--10698, 2020.

\bibitem[\protect\citeauthoryear{Zagoruyko and
  Komodakis}{2016}]{zagoruyko2016wide}
Sergey Zagoruyko and Nikos Komodakis.
\newblock Wide residual networks.
\newblock {\em British Machine Vision Conference}, 2016.

\bibitem[\protect\citeauthoryear{Zhang \bgroup \em et al.\egroup
  }{2020}]{zhang2020semantics}
Zhuosheng Zhang, Yuwei Wu, Hai Zhao, Zuchao Li, Shuailiang Zhang, Xi~Zhou, and
  Xiang Zhou.
\newblock Semantics-aware bert for language understanding.
\newblock In {\em AAAI Conference on Artificial Intelligence}, volume~34, pages
  9628--9635, 2020.

\bibitem[\protect\citeauthoryear{Zhou}{2018}]{zhou2018brief}
Zhi-Hua Zhou.
\newblock A brief introduction to weakly supervised learning.
\newblock {\em National science review}, 5(1):44--53, 2018.

\bibitem[\protect\citeauthoryear{Zhu and Goldberg}{2009}]{zhu2009introduction}
Xiaojin Zhu and Andrew~B Goldberg.
\newblock Introduction to semi-supervised learning.
\newblock {\em Synthesis lectures on artificial intelligence and machine
  learning}, 3(1):1--130, 2009.

\bibitem[\protect\citeauthoryear{Zhu \bgroup \em et al.\egroup
  }{2020}]{zhu2020incorporating}
Jinhua Zhu, Yingce Xia, Lijun Wu, Di~He, Tao Qin, Wengang Zhou, Houqiang Li,
  and Tie-Yan Liu.
\newblock Incorporating bert into neural machine translation.
\newblock {\em arXiv preprint arXiv:2002.06823}, 2020.

\end{thebibliography}
\end{document}